\documentclass[runningheads]{llncs}

\usepackage[T2A,T1]{fontenc}
\usepackage[utf8]{inputenc}
\usepackage[russian,english]{babel}
\usepackage{graphicx}
\begin{document}
\title{Endangered Languages are not Low-Resourced!}
%
%
\author{Mika Hämäläinen \orcidID{0000-0001-9315-1278}}
\authorrunning{M. Hämäläinen}
%
\institute{University of Helsinki, Finland \\
\email{mika.hamalainen@helsinki.fi}}
\maketitle              
\begin{abstract}
The term low-resourced has been tossed around in the field of natural language processing to a degree that almost any language that is not English can be called "low-resourced"; sometimes even just for the sake of making a mundane or mediocre paper appear more interesting and insightful. In a field where English is a synonym for language and low-resourced is a synonym for anything not English, calling endangered languages low-resourced is a bit of an overstatement. In this paper, I inspect the relation of the endangered with the low-resourced from my own experiences. 

\end{abstract}
\section{Introduction}

\textit{Low-resourced}\footnote{The form preferred by Dr Jack Rueter} or \textit{low-resource} language is one of those terms that has never been defined in our field, and yet it has been used very often in many publications in the past, present and future. As it is a term that is supposed to be implicitly understood without any actual thresholds for the number of speakers or the amount of data etc., it is no wonder that people use the term as they please.

Several tasks have been called low-resourced in the NLP research, for languages such as: Chinese \cite{xu2017transfer} (1.2 billion speakers), Arabic \cite{halabi2017arabic} (422 million speakers), Bengali \cite{rezaul2020deephateexplainer} (228 million speakers), Japanese \cite{yu-etal-2020-hypernymy} (126 million speakers), Vietnamese \cite{kruengkrai-etal-2020-improving} (76 million speakers), Dutch \cite{perl-etal-2020-low} (24 million speakers), Sinhala \cite{karunanayake-etal-2019-transfer} (17 million speakers) and Finnish \cite{alnajjar2019no} (5 million speakers). To make matters worse, even endangered languages have been called low-resourced \cite{55cd7fe493254a8b9d0f98323b98ab35,soisalon-soininen-granroth-wilding-2019-cross,moeller-etal-2019-improving,lim2018multilingual}. The listing is not exhaustive, but it illustrates the problem how little semantic value the term low-resourced has, given that any language can be called low-resourced.

As someone who is native in a relatively small language, Finnish, I have a very different perspective to what low-resourced means. Finnish has around 5 million speakers, and since that is around a half of the population of Tokyo (9.3 million), our language does seem small. At any case, I would never call my native language low-resourced\footnote{Unless I wanted to get a mediocre paper accepted}. Why is this, you might ask? Calling Finnish low-resourced is denying the fact that we have our own TV shows, movies, music, theater plays, novels and other cultural products in Finnish. And I am not even talking about small numbers here. Besides, Finnish is a language of education, there is no level of schooling you could not complete entirely in Finnish, from pre-school to defending your PhD. Yes, our language is small on a global scale, but we do have a whole bunch of resources we generate every day by communicating with each other!

Now, one of the main problems I find with the term low-resourced, when used about languages that have millions of speakers, is that the resources are always out there. It is, perhaps, more often than not a question of learned helplessness of a researcher. There are many ways of doing annotation projection or just gathering data in a savvy way by crawling the internet. If some NLP resource does not exist for a language, stop complaining about how low-resourced it is, get up and gather the data. Of course, there are always exceptions when gathering the data required for a large language might not be a walk in a park such as when dealing with historical data \cite{10.1093/llc/fqz024}. And it is true that even resources for non-endangered languages can be noisy \cite{befd39df758e43fb87572aa4ace5037a}. However, working with a non-endangered language does not have the same degree of problems as endangered languages might have, as I will describe later in this paper.

Quite often there are a lot of annotated resources for a majority language that are hidden on a hard drive of a researcher or published somewhere where they can be difficult to find. This leads to a situation where a language might seem low-resourced initially, but the resources are already out there, pre-annotated. They are just hidden somewhere where Google will not find them, or in the Nordic context, hidden behind a Korp user interface \cite{borin2012korp} in such a way that the resources have no value for NLP.

The main purpose of this paper is to wake people working with NLP up. If we want to continue using the term low-resourced, we had better define it as a community. Or much rather come up with a classification of how low-resourced a language is. It is about time we stopped using the term low-resourced as a fancy term to boost our papers or as an excuse for not annotating data (and releasing it for others to use).

\section{Endangered, but How Endangered?}

For the reasons that should have become evident by now, I feel that calling any endangered language low-resourced is an overstatement, as it clearly lifts a truly resource-poor language into the same league with the big players. Before continuing any further in this section, I would like to point out that much like low-resourced languages, endangered languages are not a homogeneous group. There is a huge variety in the digital resources and socio-political status these languages and their speakers have. As an example, UNESCO \cite{moseley_2010} categorizes endangered languages into 5 categories: vulnerable, definitely endangered, severely endangered, critically endangered and extinct, depending on the level of intergenerational language transmission.

In this section, I will describe some of my encounters with endangered language communities. I know that my experiences do not reflect all endangered languages, but I feel that it is important to share them to better contextualize what endangered languages can be. This is something that gets easily forgotten when doing NLP research as a rich language and culture get very easily reduced into a dataset, machine learning model and results.

I had an opportunity to visit Ufa, the capital of the Republic of Bashkortostan, Russia. The local language, known as Bashkir (bak), is rated vulnerable according to UNESCO with its more than 1.6 million speakers. While visiting the premises of Bashkir Encyclopedia\footnote{http://www.bashenc.ru/} (\foreignlanguage{russian}{\textit{Башкирская энциклопедия}}), it became evident to me that there was no lack of high-quality written knowledge in Bashkir. The number of different encyclopedias about different topics they showed us was incredible. Not to mention that they were very serious about writing the encyclopedias, according to them, they only believed in facts and numbers. The encyclopedias were not thus just mere translations of existing ones, but rather their own independent work of science. I was also told about TV channels broadcasting TV shows in Bashkir, which while interesting, is not that surprising given the number of speakers.

FU-Lab\footnote{https://fu-lab.ru/} welcomed me for a research visit in the capital of the Komi Republic, Russia, Syktyvkar. While there are several Komi languages, our discussions were mainly related to Komi-Zyrian (kpv), a language marked as definitely endangered in the classification of UNESCO with as few as 217000 native speakers. Seeing the work conducted in FU-Lab in action, one can say that Komi-Zyrian has a surprising amount of digital resources. They actively develop constraint grammar based disambiguators and contribute to the morphological FSTs (finite-state transducers). In addition to that they compile text and audio corpora for Komi (see \cite{fedina_komi_corpus,fedina_mediateka,fedina_corpus_komi}) and have many dictionaries available in a digital format. An interesting decision by the Komi Republic that could eventually lead into functional Komi-Russian machine translation is the fact that all Russian law texts are translated into Komi. At FU-Lab, they have ensured that the translations remain parallel to the original Russian law texts, which should make machine translation easy in the future.

Our system Ve\textsuperscript{$\prime$}rdd \cite{alnajjar2020ve} was the reason I got an opportunity to visit the Sami Culture Center Sajos\footnote{http://www.sajos.fi/} in Inari, Finland to collaborate with two Skolt Sami dictionary editors. Skolt Sami (sms) is a severely endangered language with only 300 native speakers according to UNESCO. Despite the low number of speakers, they had the presentations of the Sami cultural event simultaneously interpreted from Skolt Sami to Finnish and from other Sami languages to Skolt Sami by professional interprets. Thanks to Rueter's continuous efforts for the digital revitalization of the language, Skolt Sami has an extensive digital multilingual dictionary \cite{263590f5c0614385a3fe982ba43fd84e} and FST morphology \cite{rueter2020fst}. The situtaion of Skolt Sami is fortunate in the sense that it is one of many Sami languages. The Sami Parliament has established Sámi Giellagáldu to do work on language norms and terminology for various Sami languages including Skolt Sami.

As a summary, it is important to understand that the term endangered language is complicated as well in terms of the linguistic resources available for NLP tasks. Some endagered languages may have a surprising amount of resources for some specific NLP tasks, while others may not have digital resources at all. For more anecdotes, I strongly recommend reading Rueter's personal experiences of everyday situations in Erzya \cite{rueter2013erzya}.

\section{The Underappreciated Problem of Being Endangered}

There have been several papers trying to tackle low-resourced tasks either by simulating a resource-poor scenario in a high-resourced language or by having limited resources for a high-resourced language \cite{gu2018universal,tiedemann2020tatoeba,126be25f4261423faaf3b829f458ba19}. It is important to remember that while these efforts are of value in many domains, they might not be directly applicable as such for endangered languages.

The issues you might face with an endangered language start from the very low-level: character encoding. I am not referring to any custom or local encoding, but to Unicode, the encoding we know and love. Unicode is not at all as uncomplicated when we are dealing with smaller languages. One of the problems is that Unicode has multiple ways of encoding one character and that there are similar looking, but not quite the same characters.

\begin{figure}[h]
  \centering
  \includegraphics[width=\linewidth]{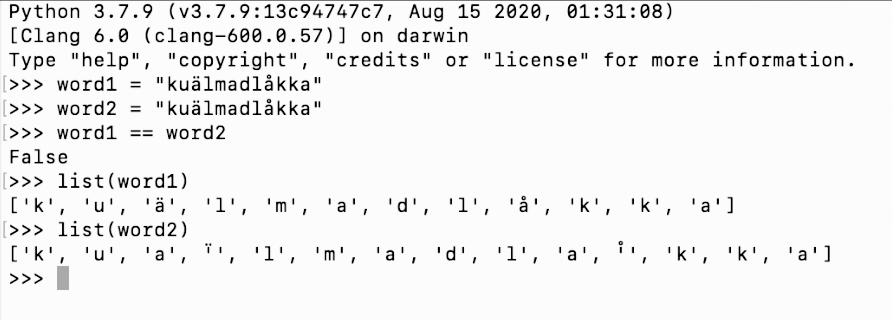}
  \caption{A typical situation with Skolt Sami data.}
  \label{skolt_mess}
\end{figure}

Figure \ref{skolt_mess} illustrates the aforementioned problem. The word \textit{kuälmadlåkka} seems to be written similarly in \textit{word1} and \textit{word2}. However, they are not identical, as seen when they are split into characters. This type of an issue is not as common in high-resourced languages and it might go unnoticed for the unwary researcher. These issues are present with some endangered languages because of a multitude of reasons such as a simple lack of a suitable keyboard layout, lack of a standardized orthography or a change in orthography. Although, for practical reasons, you might see people writing words in a non-standard way even in larger languages such as writing \textit{paral.lel} instead of \textit{ paral·lel} in Catalan or \textit{ca} instead of \textit{ça} in French, however the reason for these non-standard ways of writing are different than in the Skolt Sami example. At any rate, misspellings are common in texts written in an endangered language as well, as reported for North Sami \cite{antonsen2013callinmeattahusaid}.

\begin{figure}[h]
  \centering
  \includegraphics[width=6cm]{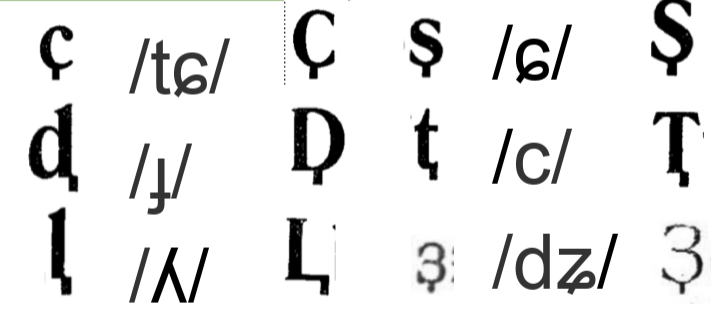}
  \caption{Latinitsa letters missing from Unicode}
  \label{latinitsa}
\end{figure}

Sometimes not all important characters are in Unicode as seen in Figure \ref{latinitsa}. Such is the case with the writing system of Komi-Zyrian that was in place before the Cyrillics, \foreignlanguage{russian}{латиница} (latinitsa). There has been a long-running effort \cite{rueterkomi} in getting the missing characters into the Unicode standard, but so far this effort has been futile. This means that there is no consistent way of encoding historical Komi texts written in this script.

When we write anything in a big language, it is usually clear what is right and wrong. An extensive part of any educational system goes into teaching people to write in a normative way. Much of this normative writing is something we who are native in a non-endangered language just learn by being exposed to normative language in various places such as books and news. Norms are usually established and maintained by a language institute. In Finland, this practice is called \textit{kielenhuolto} (language maintenance), a metaphor that makes any non-normative language sound broken like any machine that is in need for maintenance. Even though the "correct" way of using a language is internalized to different levels by different people, this practice will be reflected in any data produced in a high-resourced language. 

The tradition of seeking for a correct language is called prescriptive linguistics, where the focus is on how a language should be. Sometimes this prescriptive ideology is absent or the prescriptive rules are not internalized by the people native in an endangered language due to various reasons such as a limited access to education in one's own language. This leads to a situation where we cannot expect endangered language data to be "correct" in the prescriptive sense, but it might be more reflective of how people actually use the language. This still does not mean that the data would be consistently deviant of some norm. Just as with bigger languages, endangered languages, even as small as Skolt Sami have different dialects \cite{bb1983fb55f4444eb483e987c1f664a5}.

With the recent edition of the Skolt Sami dictionary \cite{skolt_dictionary} there was some discussion about certain words such as why \textit{Amerikk} (America) and \textit{ankerias} (eel) were used instead of older words \textit{Ä\textsuperscript{$\prime$}mmrikk} and \textit{aŋŋerias}. Questions like these are related to whether a language should be documented as it should be or as it is. A highly prescriptive dataset in an endangered language might thus mean no applicability to real world data as people using the language might not be aware of the norms. In the case of Skolt Sami, I would be surprised if all speakers were aware of the rules, as Sámi Giellagáldu publishes their latest recommendations in their blog, making such recommendations difficult to consult.

Endangered language data is also more prone to containing mistakes beyond encoding and lack of norms. In my work, I have noticed several mistakes in different resources such as XML dictionaries \cite{55cd7fe493254a8b9d0f98323b98ab35}. I would not point fingers and call anyone's work bad, as mistakes do happen, especially when there are several people working on the resources during different times. The reason why I believe that there are more mistakes in endangered language resources is the simple fact that there are fewer people inspecting them and pointing out errors. It is very common to become blind to one's own work, and spotting mistakes requires external inspection.

\section{Do Only Rules Rule?}

One of the things that divide people in NLP is rules versus neural networks. Why would you write rules for an endangered language if neural networks work for a low-resourced Hindi? As we have seen in this paper, the problems endangered languages have are not the same as just about any "low-resourced" language would have. But at the same time I am facing the ideology that only rules can be used to model endangered languages. I, myself, don't believe that either rules or neural networks are the answer. The optimal solution is probably somewhere in the middle ground. 

Rule-based methods are continuously developed for various endangered and extinct languages \cite{schmirler2019modelling,pirinen2019building,sahala-etal-2020-babyfst}. Sometimes, due to the lack of resources, rules are the only viable way of dealing with these languages. I believe that here rules serve for a more important role than mere engineering of an NLP system. Many endangered languages are under-documented, and machine readable rules serve for language documentation purposes as well. They need to capture something meaningful about the language being described in order for them to work to begin with. From this perspective, I think that rule-based systems are not only valuable from the point of view of NLP but also from the point of view of linguistic research. Only machine readable rules let you test out your linguistic hypotheses extensively. 

Rules are good in the sense that they can be fixed easily, and it is possible to reach to a high accuracy with them. This is useful when building systems like spell checkers and language learning tools, as the correctness of these tools is of utmost importance. However, rules can only go so far. An FST, no matter how extensive, is never going to contain all the words in its vocabulary, for example. For this reason, neural networks are useful, as they can produce output even for new input that was not present in the training data. However, usually accuracy is important in the context of endangered languages as many of the NLP tools are built for practical purposes and for the benefit of the language community.

Rules can get things right, but their limits can be reached easily. Neural networks can go beyond a predefined set of rules, but they are more prone to producing incorrect results as well. I think that the two different approaches should be used together. Rules can be used to generate training data for neural networks, something I call \textit{fake it till you make it approach}, or they can be used to filter out low-quality samples from a training dataset. This way, rules can be used in the training process. Because rules can be easily fixed, I would pipeline rules with neural networks. Whatever rules cannot cover, a neural network can handle, and if the system produces wrong output, the rule-based method can always be fixed.

I think that synthetic data generation is still an under-studied way of building NLP tools for endangered languages. FSTs are a good way of achieving this as they can be used both for generation and analysis. We have reached to rather good results for some languages with FST generated data and you can expect to see neural models integrated with UralicNLP \cite{uralicnlp_2019}\footnote{https://github.com/mikahama/uralicNLP} as a backup for failing FSTs in the near future.

\section{Conclusions}

The term low-resourced is truly a complicated one and it makes NLP research conducted for endangered languages difficult to get published in the bigger ACL venues. Work with endangered languages is not as state-of-the-art driven as it is usually to be expected from NLP papers in bigger venues. Instead such a work is more practical, typically involving producing tools and resources for the benefit of the language community.

Any work with endangered languages includes ethical considerations. I have always been puzzled by the fact that in the world of NLP research not releasing one's code, data and models is considered acceptable practice. With endangered languages not releasing the resources produced is even more severe as such a behavior may be interpreted more as a cultural and linguistic appropriation of a vulnerable group of people purely for the sake of academic merit.

Endangered languages pose very different types of challenges for NLP research and they have very different amounts and types of resources available. Some languages have quite advanced NLP tools in place thanks to altruistic research endeavors and active community members, while others do not have anything. As there is such a huge variation within the group of endangered languages, grouping them together with anything "low-resourced" from Chinese to Finnish\footnote{I still don't think Finnish is low-resourced} is very misleading.

I have shared my personal experiences working with NLP for endangered languages and working with people who are native in some of them. A lot has been left unsaid, and I do know that there are a whole lot of languages out there that are dealing with very different issues than what I have described in this paper (c.f. \cite{hammarstrom2010status}). The main purpose of my descriptions has been to show people what type of problems one can encounter when conducing this type of a research. I am honored for having had this possibility of seeing NLP beyond large languages.

\section*{Acknowledgments}

I would like to thank Jack Rueter, who was kind enough to supervise my PhD thesis even though it ended up being unrelated to Jack's research interests. Without him, I would have never been exposed to endangered languages nor would have I ever had the chance of traveling to the places where such languages are spoken. Also, without his fantastic resources like FSTs \cite{rueter-etal-2020-open,71debe3f0608471284bc2419a2f392de}, dictionaries \cite{rueter2019xml}, corpora \cite{12fa63d7f4bf41bda1896960543de4b6} and Universal Dependencies treebanks \cite{rueter-tyers-2018-towards,rueter-etal-2020-questions} NLP would be a distant dream for many endangered Uralic languages.

%
%
%
 \bibliographystyle{splncs04}
 \bibliography{mybibliography}
\end{document}